\documentclass[letterpaper, 10 pt, conference]{ieeeconf}  

\IEEEoverridecommandlockouts                           
\usepackage{graphicx}      	
\usepackage{listings}	   	
\usepackage{float}
					   	
\usepackage{graphics} 

\title{\LARGE \bf
Towards Automatic Migration of ROS Components from\\Software to Hardware
\\[5mm]\em\normalsize DSLRob 2013 --- Work-in-progress
}

\author{Anders Blaabjerg Lange, Ulrik Pagh Schultz and Anders Stengaard Soerensen
\thanks{A. B. Lange, U. P. Schultz and A. S. Soerensen are with the Maersk McKinney Moeller Institute, University of Southern Denmark, Odense, Denmark
(e-mail: $\lbrace$anlan, ups, anss$\rbrace$@mmmi.sdu.dk)}
}

\begin{document}

\maketitle
\thispagestyle{empty}
\pagestyle{empty}


\section{Introduction}

The use of the ROS middleware is a growing trend in 
robotics in general, in particular in experimental branches of robotics 
such as modular robotics, fields robotics, and the vast area of cyber-physical systems (for example applied to welfare technology).
Our main area of interest is in experimental robotics and cyber-physical systems. When building ``robot controllers'' for the 
aforementioned systems there are numerous suitable technological platforms.  
Given specific requirements we can choose an appropriate standardized 
approach, for example emphasizing flexibility and ease of development 
by using a generic middleware --- such as ROS --- or emphasizing real-time 
performance and direct hardware access by using approaches based on 
dedicated, embedded hardware.  So far ROS and hard real-time 
embedded systems have however not been easily uniteable while retaining the same overall communication and processing methodology at all levels. 

In this paper we present an approach aimed at tackling the schism
between high-level, flexible software and low-level, real-time
software.  The key idea of our approach is to enable software
components written for a high-level publish-subscribe software
architecture to be automatically migrated to a dedicated hardware
architecture implemented using programmable logic.  Our approach is 
based on the Unity framework, a unified software/hardware framework 
based on FPGAs for quickly interfacing high-level software to low-level 
robotics hardware. The vision of Unity is to enable non-expert 
users to build high-quality interface and control systems using FPGAs 
and to interface them to high-level software frameworks, thereby 
providing a framework for speeding up and increasing innovation in 
experimental robotics.  This paper presents the overall vision and the 
initial work on the implementation of an architecture supporting a 
generative approach, based on a declarative specification of how 
software components are mapped to a hardware architecture; the actual 
language design is left as future work.

\section{Context: Unity and FPGAs}

The traditional approach to building a control system in experimental
robotics is mainly based on microcontrollers (MCU's) and PC's. This
approach has numerous advantages, mainly: (1)~developers are familiar
with the programming methodology; (2)~good tools, libraries and
frameworks from commercial vendors and the open-source community; and
(3)~the availability of cheap and simple MCU-based systems like the
Arduino, as well as more powerfull ARM based systems.  Despite the advantages of this approach, there are also
inherent limitations to the sequential-style processing and fixed
hardware (HW) architecture, which can significantly limit reuse of HW
as well as real-time capabilities, design freedom and flexibility.

We prefer FPGAs and hybrid FPGA-MCU SoC systems over pure MCUs: we 
find FPGAs superior to MCUs in many performance areas relevant to 
experimental robotics, except for price and library support. FPGAs can 
provide deterministic hard real-time performance no matter the 
complexity or scale of the implemented algorithms~\cite{fernandes, 
pordel, toscher}. On an FPGA the architecture is designed by the 
developer, providing increased flexibility that can reduce the need 
for costly software abstractions on higher levels \cite{falsig1, 
falsig4, falsig3, falsig2} and reduce or eliminate the need for 
external support logic. FPGAs are however not commonly used; we 
believe the reason to be partly historical: people stick to 
technologies they know. Moreover, FPGAs suffer from a lack of good, 
open-source, vendor-independent HDL-component libraries suited for 
robotics, and a high degree of complexity associated with FPGA 
programming, caused partly by complex tools and a different 
programming methodology compared to the traditional Von-Neumann style.

\begin{figure*}[t]
\begin{center}
\center\includegraphics[width=15.3cm]{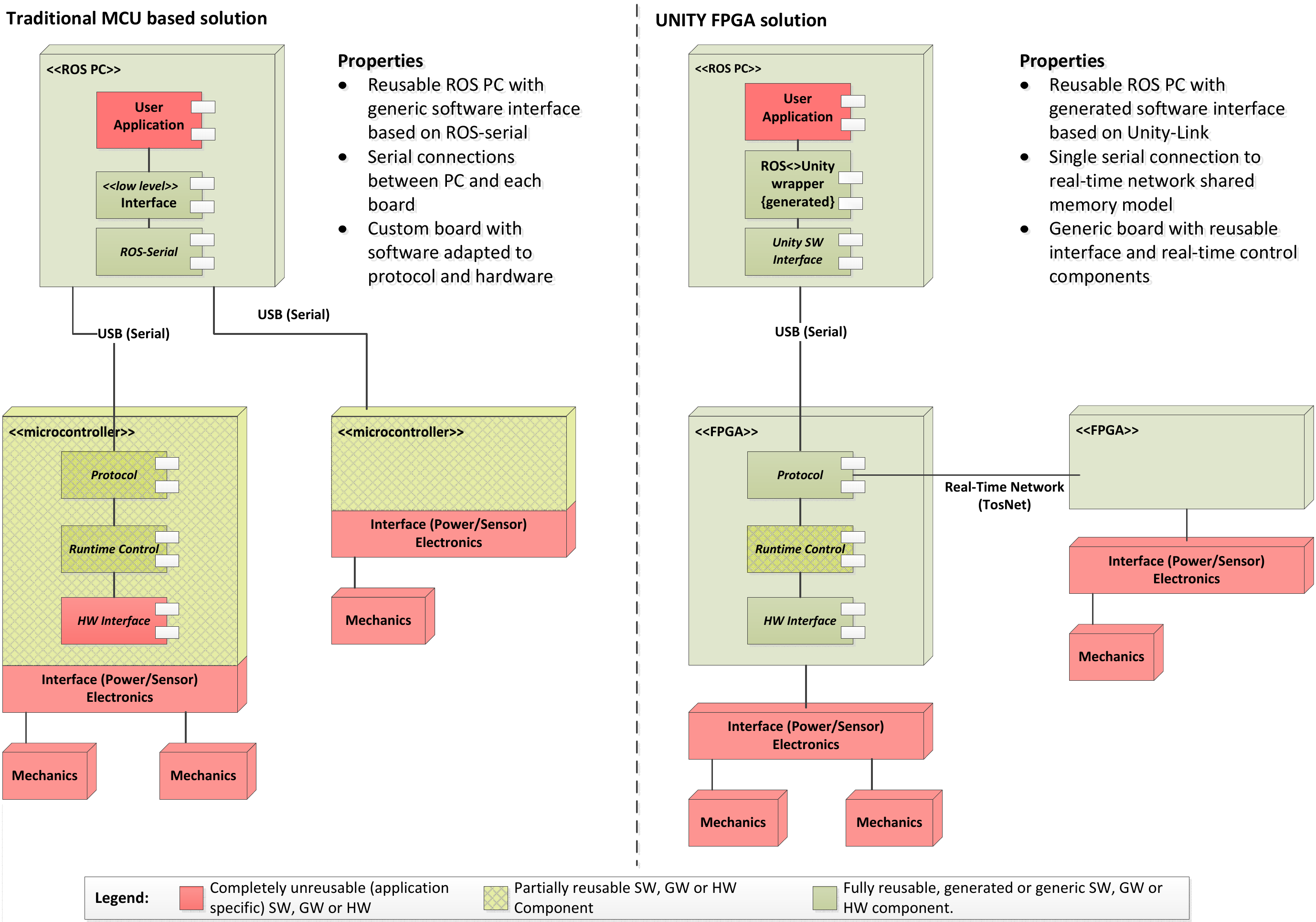}
\caption{Unity Link compared to a traditional MCU-based architecture (UML 2.0 component diagram notation)}
\label{fig:unity-vs-mcu}
\end{center}
\end{figure*}

We have proposed the Unity framework as a means to facilitating
FPGA-based development for experimental
robotics~\cite{sdir8,iros13}. Unity is an open-source framework
consisting of reference HW designs, gateware (GW, VHDL) and SW
libraries, all targeted at providing a complete framework for easy
development, with standard cases covered by model-based code
generation of all the necessary FPGA GW and PC SW needed to interface
electronics with a high-level software framework. The Unity framework
is a work-in-progress: The modular HW designs include single nodes,
distributed nodes, sensor interfaces and generic motor controllers. On
the GW side we have a growing library of VHDL modules, including
servo- and brushless DC motor controllers, a real-time network based
on a shared memory model, a complete FPGA-based real-time operating
system~\cite{hartos}, as well as a modular and reconfigurable FPGA-PC
interface called Unity-Link~\cite{iros13}. The use of Unity compared
to a traditional MCU-based approach, exemplified with a PC connected
to low-level hardware using ROS-serial, is illustrated in
Figure~\ref{fig:unity-vs-mcu}. 
We believe that a generic FPGA or FPGA-MCU SoC based module will be 
more flexible and therefore more easily reused for various tasks, 
compared to a standard off-the-shelf MCU system, since the various 
hardware interfaces needed are decoupled from (i.e., not locked to) 
specific pin locations, and therefore virtually only the pin count 
limits the number and types of interfaces that are possible when using 
programmable logic. Unity is an evolution of the 
TosNet framework, which is the basis for the real-time network and 
other specific components~\cite{falsig1,falsig4,falsig3,falsig2}.

\section{Automatic Migration of ROS Components to FPGAs}

\begin{figure*}[t]
\begin{center}
\center\includegraphics[width=14.8cm]{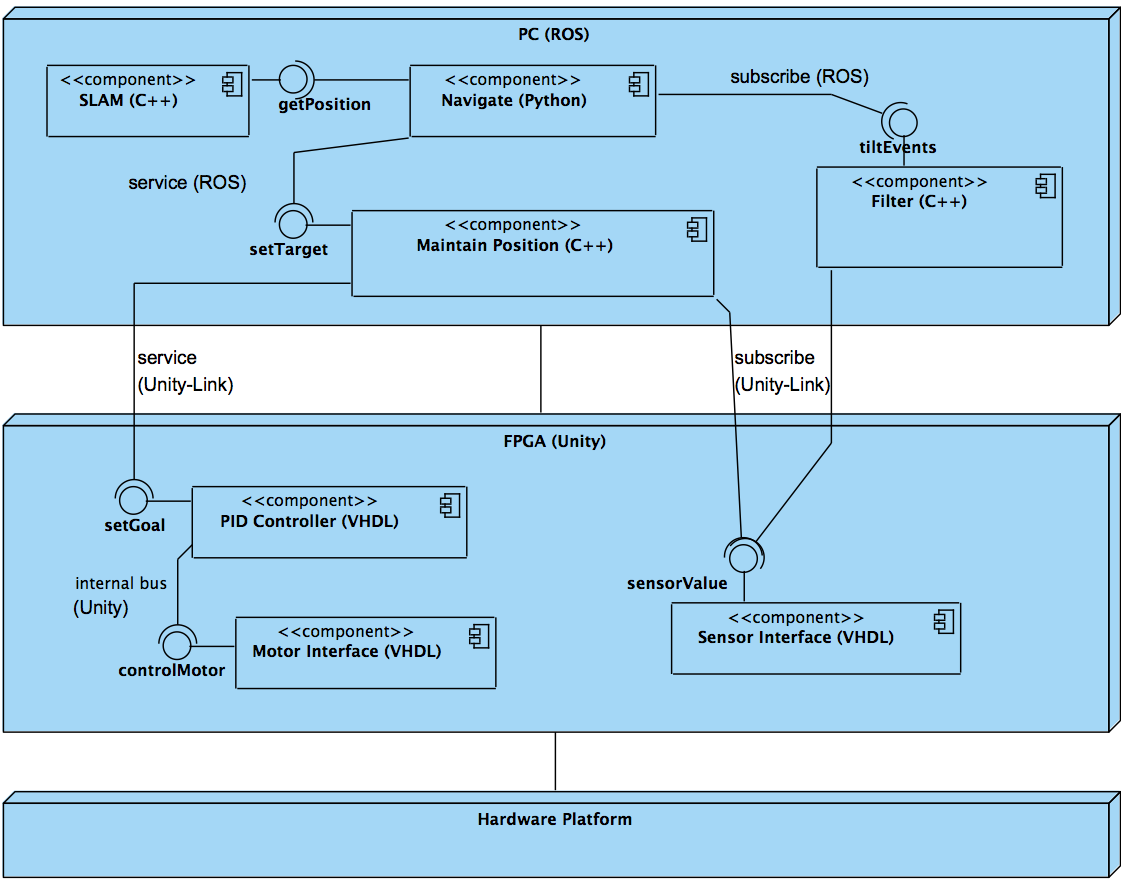}
\caption{Example: Robot software before migration of selected
  components, part of the control loop is implemented in ROS. (UML 2.0)}
\label{fig:migration-before}
\end{center}
\end{figure*}

\begin{figure*}[t]
\begin{center}
\center\includegraphics[width=14.8cm]{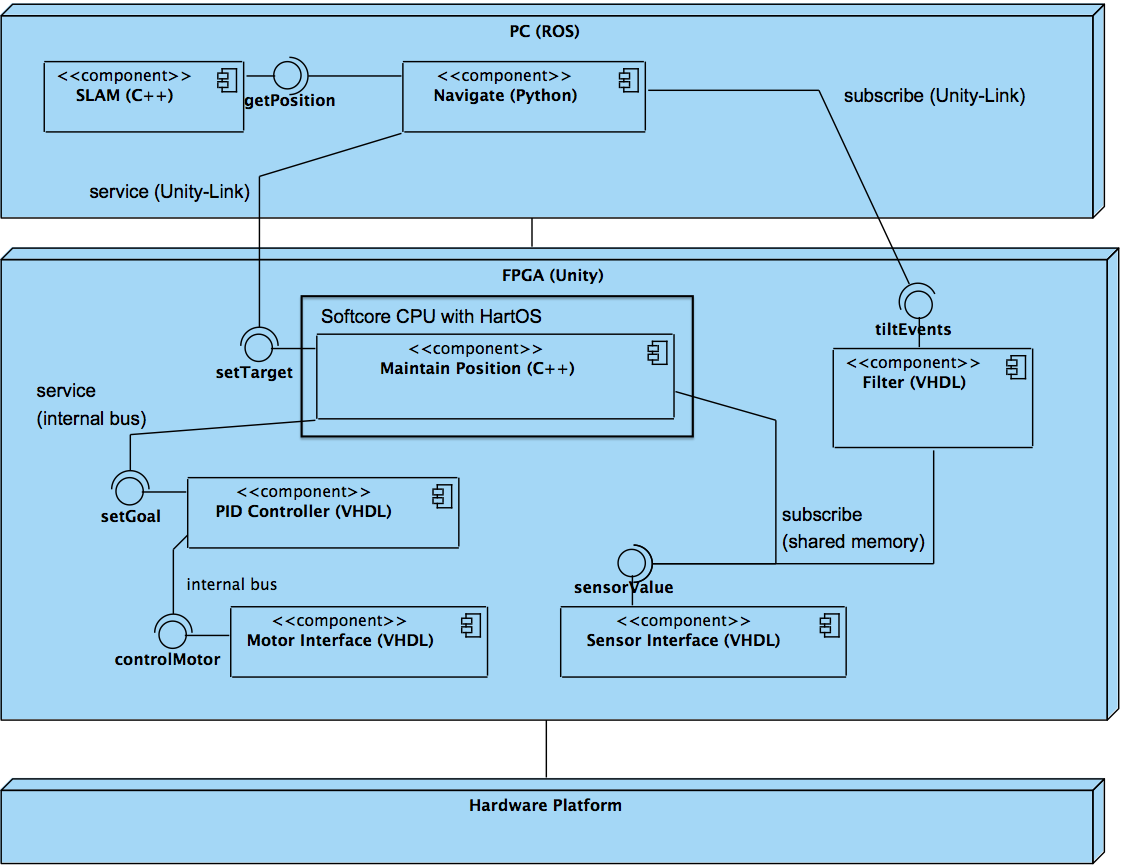}
\caption{Example: Robot software after migration of selected
  components, the ``maintain position'' and ``filter'' components have
  been migrated to the FPGA. (UML 2.0)}
\label{fig:migration-after}
\end{center}
\end{figure*}

We are currently investigating the idea of automatically migrating
networks of ROS components\footnote{Throughout this paper we
  consistently use the term ``ROS component'' to refer to ROS nodes:
  we believe our approach is applicable to other component-based
  middlewares as well, and hence prefer the technology-independent
  term ``component.''} to our FPGA-based architecture.  The Unity
framework already provides a standardized platform on which gateware
components can be interconnected, and Unity-Link provides automatic
integration on ROS components with gateware components using a
publish-subscribe infrastructure~\cite{iros13}.  There is however no
support for migrating a ROS component, or a set of ROS components,
from the PC to the FPGA, without completely reimplementing the
functionality of each of the components, and furthermore using
the Unity framework to connect them internally on the FPGA.  Note that
we are not concerned with dynamic migration: we simply want to make it
easy for the developer to statically change the deployment of functionality
between the flexible PC platform and the real-time FPGA platform.

We propose that migration of a given ROS component from the PC-based
platform to the FPGA can be done by recompiling the component to run
on either a softcore or hard-IP CPU embedded in the FPGA.  The HartOS real-time operating system~\cite{hartos}
will be used to execute the threads of the component and to handle
external events.  A substrate that provides the ROS API and a few
selected parts of the standard POSIX API\footnote{Only a
  small subset of the POSIX API will be relevant, as well as feasible for a processing system utilizing the HartOS kernel.  We assume our
  approach is primarily relevant for ROS components having a fairly
  small amount of interaction with the operating system.} will be used
on the embedded CPU, enabling a ROS component e.g.\ implemented in
C$++$ to execute on the CPU after a simple recompilation.
Publish-subscribe messages can be routed between the CPU and a PC running ROS using Unity-Link. A high performance hard-IP CPU, like e.g.\ the dual-core ARM-A9 in a Xilinx Zynq device, could as a second option also run a full linux system with ROS, and thereby support native (non-recompiled) ROS components.  By providing the same memory-mapped publish/subscribe and service-call IP interfaces on both the small softcore and Hard-IP CPU's, no matter the software environment executed on them, Unity will allow both high and low performance processors, and PC's using Unity-link, to communicate with GW components directly utilizing ROS' own communication paradigm, thereby enabling easy migration between
execution paradigms.

A set of ROS components that communicate using publish-subscribe can
similarly be migrated to the FPGA.  Each component is placed on a
softcore CPU, depending on the performance requirements they can be
placed on the same or different CPUs.  If they are placed on the same
CPU, HartOS is used for scheduling CPU-time between the components,
and communication can be performed directly between the components
(taking care to preserve communication semantics).  If components are
placed on different CPUs, a shared memory component is used
to propagate publish/subscribe messages between the nodes: each topic
uses a specific address in the shared memory, enabling a complete
decoupling of the execution of publishers and subscribers.  Service
calls can be handled similarly, however rather than using a shared
memory, a generic address-data bus can be used to provide a
point-to-point connection between components that need to communicate.

As an example, consider a first revision of the robot software
architecture for a two-wheeled balancing robot shown in
Fig.~\ref{fig:migration-before}.  Low-level control and hardware
interfacing is done in the FPGA using the Unity framework, and
consists of low-level hardware interface components and a generic PID
controller.  Unity-Link connects low-level control and sensor
interfaces to ROS using publish/subscribe and service calls.
High-level control is implemented in ROS, and concerns navigation,
movement, and balancing of the robot.  Real-time operation of the
``maintain position'' component is ensured by using a suitably fast
PC. Now assume that --- although initial experiments showed that this
worked fine --- after experimenting with the robot in a realistic
scenario it is found that control is unstable because real-time
deadlines are sometimes missed.  To solve the issue using our
approach, the ``maintain position'' component is moved to a softcore
CPU on the FPGA, as illustrated in Fig~\ref{fig:migration-after}.
Moreover, due to the use of standard interfaces that are interoperable
between ROS and the FPGA, the software filter component is
transparently replaced by a functionally equivalent gateware component from the Unity library.  All components on the FPGA execute in hard real-time, making control of the robot predictable.

\section{Discussion and Status}

The migration is intended to be automatic, in the sense that given a
declarative specification of how a set of ROS components should be
mapped to a real-time architecture, our system will generate substrate
code, configuration files, and VHDL components such that the ROS
components can be directly recompiled to run on the FPGA.  This
declarative specification will thus need to model which components are
to be deployed to which softcore CPU, how much time is to be assigned
to each thread of each component, and how the components are to
communicate with each other and with the rest of the ROS system.

We are currently extending the Unity framework to support multiple ROS
components executing and communicating in real-time on one or more
softcore CPUs on one or more FPGAs connected by a real-time network.  Once
this framework is complete, we will augment it with a model-based code
generator than can automatically generate the complete set of code
artifacts needed to support the execution of the ROS components on the
FPGA.  We expect that the task of implementing the framework and
corresponding generator is significantly reduced by building on top of
the standardized Unity architecture and using Unity Link to interface
the FPGA-based ROS components to the rest of the ROS system.





\end{document}